\def\year{2021}\relax
\documentclass[letterpaper]{article} 
\usepackage{aaai21}  
\usepackage{times}  
\usepackage{helvet} 
\usepackage{courier}  
\usepackage[hyphens]{url}  
\usepackage{graphicx} 

\usepackage{natbib}  
\usepackage{caption} 
\usepackage{multirow}
\usepackage{amsmath}
\usepackage{textcomp}
\title{Deeplite Neutrino\texttrademark: \\ An End-to-End Framework for Constrained Deep Learning Model Optimization}
\author{
    Anush Sankaran$^1$, Olivier Mastropietro$^1$, Ehsan Saboori$^1$, Yasser Idris$^1$, Davis Sawyer$^1$, MohammadHossein AskariHemmat$^{1, 2}$, Ghouthi Boukli Hacene$^{1, 3}$ \\
}
\affiliations{
    $^1$Deeplite Inc., Montreal \\ $^2$ Ecole Polytechnique, Montreal, \\ $^3$ MILA, Montreal
}

\begin{document}

\maketitle

\begin{abstract}
Designing deep learning-based solutions is becoming a race for training deeper models with a greater number of layers.
While a large-size deeper model could provide competitive accuracy, it creates a lot of logistical challenges and unreasonable resource requirements during development and deployment. This has been one of the key reasons for deep learning models not being excessively used in various production environments, especially in edge devices. There is an immediate requirement for optimizing and compressing these deep learning models, to enable on-device intelligence.
In this research, we introduce a black-box framework, Deeplite Neutrino\texttrademark~for production-ready optimization of deep learning models. The framework provides an easy mechanism for the end-users to provide constraints such as a tolerable drop in accuracy or target size of the optimized models, to guide the whole optimization process. The framework is easy to include in an existing production pipeline and is available as a Python Package, supporting PyTorch and Tensorflow libraries. The optimization performance of the framework is shown across multiple benchmark datasets and popular deep learning models. Further, the framework is currently used in production and the results and testimonials from several clients are summarized.
\end{abstract}

\section{Introduction}
Deep learning has been one of the most intrusive technologies of the 21st century, having revolutionized businesses across multiple industries. From building better gaming opponents to translating languages in real-time, to the detailed understanding of large volumes of images and videos, deep learning has enabled us to achieve automation in different applications. However, deep learning is now a race for the ability to build deeper and larger models to produce better results. Recent models such as BiT-M from Google~\cite{kolesnikov2019big} with 928 million parameters, Megatron-LM from NVIDIA~\cite{shoeybi2019megatron} with 8.3 billion parameters, Turing-NLG from Microsoft~\cite{rasley2020deepspeed} with 17 billion parameters, and GPT-3 from OpenAI~\cite{brown2020language} with 175 billion parameters show the unprecedented growth in the size of deep neural network (DNN) architectures.

This explosive growth has led to the primary challenge of the democratization of deep learning. Training such huge models would require vast computing powers with supercomputers, which is not accessible to all. For example, the latest GPT-3 model with over 350GB in memory size costs over \$12 million dollars to train using specialized super computers\footnote{https://venturebeat.com/2020/06/01/ai-machine-learning-openai-gpt-3-size-isnt-everything/}. Such a computing infrastructure is not available to everyone and is globally not affordable by all deep learning startups and researchers. 

There are other implications in training huge DNN models such as energy and power consumption. Strubell et al.~\cite{strubell2019energy} studied that the compute required to train large-scale DNN models produces carbon-dioxide emissions equivalent to five times the lifetime emissions of an average American car. They also showed that the annual power consumption of cloud computing giants such as Amazon AWS, Google, or Microsoft, is equivalent to the annual power consumption of the United States.
Additionally, according to a recent Gartner survey, as of 2020, there are more than six billion edge devices\footnote{https://www.gartner.com/en/newsroom/press-releases/2019-08-29-gartner-says-5-8-billion-enterprise-and-automotive-io} and current state-of-the-art DNN models are not equipped to be deployed directly on edge devices due to challenges in their memory requirements.


Our objective is to optimize such DNN model architectures without a reduction in accuracy, as step progress towards enabling them to be directly deployed in edge devices. The idea behind model optimization is under the presumption that DNN architectures are over-parameterized. Optimization reduces the number of parameters of the large DNN model while improving the performance of the model in metrics such as computational cost, inference time, and energy consumed. This leads to the primary and the most important research question, ``Can smaller models with fewer parameters, achieve an accuracy performance equivalent to a deeper model with a larger number of parameters?"

\begin{figure*}[!ht]
	\includegraphics[width=.95\textwidth]{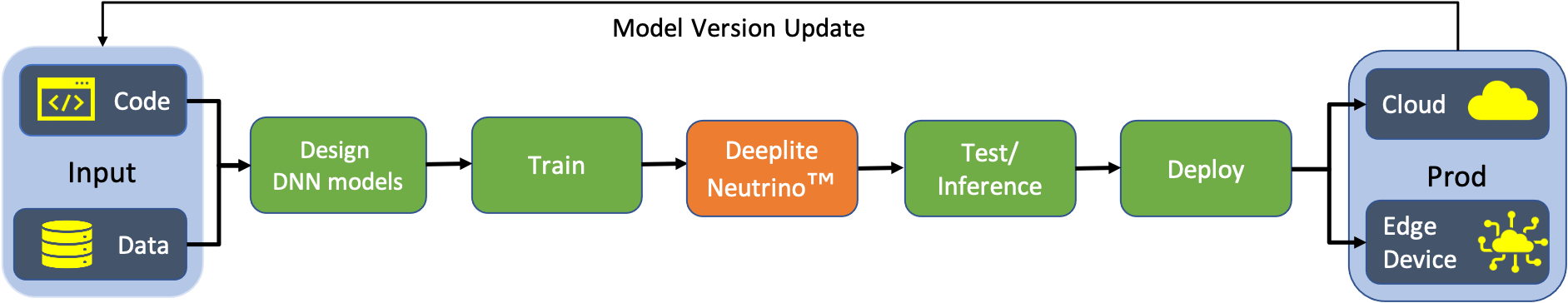}
	\caption{The typical development life cycle of a deep neural network (DNN) model. The proposed black-box framework for model optimization, Neutrino, can be seamlessly integrated into the development life cycle at minimal cost.}
	\label{fig:overview1}
\end{figure*}

\noindent \textbf{Challenges of Model Optimization in Production}

In the research community, there are some popular approaches such as model pruning, model quantization, and model decomposition to achieve model compression. However, there are a lot of challenges in consuming research oriented techniques in production.

\begin{enumerate}
    \item \textbf{Democratization of DNN Optimization:} Training and optimization of DNN architecture is currently unaffordable and requires super-computing infrastructure. How could we make a production-ready optimization framework that is consumable and affordable by everyone?
    \item \textbf{Multiple Metrics to Optimize:} There are multiple metrics to optimize such as (i) the number of parameters, (ii) model memory size, (iii) inference time,  (iv) computational cost in terms of FLOPs/ MACs, or (v) energy consumption. It is challenging to optimize in parallel multiple metrics of optimization.
    \item \textbf{Constrained Optimization}: Applications may require optimization to focus on certain metrics while trading off on other metrics. For example, real-time systems would require the inference time to be low while low-memory edge devices would focus on model memory size reduction. How would we guide the model optimization to favor certain metrics over others?
    \item \textbf{Hardware support}: The generic implementation in popular libraries such as PyTorch and Tensorflow does not support certain methods of model compression. Also, the model compilation and the device hardware-specific execution of the optimized model is challenging. While most of the techniques are targeted towards GPU, how could we optimize DNN architectures for specialized hardware?
    \item \textbf{Black-box Framework}: The end-users' usability and simplicity is a key requirement for consuming optimization in production pipelines. There is a big need for a black-box optimization framework, where the end-user could easily provide the trained model, the dataset, and constraints for optimization, while not be troubled with the nuances of implementation and execution.
    \item \textbf{Research Papers to Production:} Often, research papers aims at finding a highly optimized model which retains the accuracy of the original model, while the cost involved in optimization or searching for the optimized model is considered secondary. However, in production systems, the cost and the time incurred in optimizing the original model are equally important. Unstructured weight optimization is only realistic in some ideal theoretical hardware. A production-ready framework should generalize the optimization approach across a wide variety of architectures and hardware. 
\end{enumerate}

In this research paper, we introduce Neutrino~\footnote{In this paper, Neutrino~refers to Deeplite Neutrino\texttrademark}~\footnote{\url{https://www.deeplite.ai/index.html#neutrino}}, a lights-out DNN model optimization framework guided by the end-users' constraints and requirements. A typical continuous development life cycle of a DNN model is shown in Figure~\ref{fig:overview1}. The proposed Neutrino can be seamlessly and smoothly integrated into any development and deployment pipeline. The framework consumes a pre-trained DNN model, with the original train-test split data as input, in addition to optimization requirements from the end-user. Neutrino produces the optimized model that can be further used for inference either in a cloud environment or could be directly deployed on the edge device.  Neutrino builds a symphony of different model optimization and acceleration techniques. This research paper focuses on the part of constrained optimization technique used in the framework and the successful results obtained on various public benchmark datasets and popular models. Neutrino framework is distributed as Python PyPI library, with support for PyTorch~\cite{NEURIPS2019_9015} and early support for Tensorflow~\cite{tensorflow2015-whitepaper} library.

The rest of the paper is organized as follows; Section 2 provides background literature of various model optimization techniques. Section 3 explains the architecture of the proposed optimization framework. Section 4 details the experimental results obtained on various benchmark datasets and popular DNN architectures. Section 5 presents the business impact and use-cases of the proposed framework, along with the development details. Section 6 summarizes our efforts with some short-term and long-term future goals.
\section{Background Literature}

The different methods explored in the literature for DNN model optimization aims to reduce the number of parameters in the model. These techniques can be broadly grouped into three schools of thought: (1) weight pruning, (2) architecture search, and (3) weight decomposition.

\subsection{Weight Pruning}

The redundant parameters of the model that do not contribute to the effective output are pruned, resulting in a smaller model with fewer parameters. Column and structured shape pruning introduce non-zero weight values, while the channel and layer pruning reduce the size of the model.
Weight pruning in DNN architectures is a well-researched topic with a set of comprehensive survey reports~\cite{choudhary2020comprehensive, liu2020pruning, cheng2017survey}. Liu et al.~\cite{liu2020autocompress} proposed \textit{AutoCompress}, an automated experience-guided heuristic search technique to achieve extreme compression rates. Ren et al.~\cite{ren2020darb} proposed a density-adaptive regular-block (DARB) pruning technique to perform pruning at a channel row-level. Most of these techniques perform post-training pruning while Wang et al.~\cite{wang2020pruning} proposed a method for pruning a DNN architecture from scratch. They showed that comparable accuracy on models is achieved with similar computational budgets as the post-training pruning methods.

\subsection{Architecture Search}
Architecture search finds a surrogate model, from the set space of all possible DNN architectures, such that the surrogate (or student) model is much smaller with similar performance as the original model. Thus, model optimization is formulated as a learning or heuristic-driven search problems such as knowledge distillation~\cite{luo2016face, phuong2019towards, changyong2019knowledge}, Guided Network Architecture Search~\cite{kang2020towards}, or AutoML~\cite{he2018amc}, or meta learning~\cite{bai2019few}.

One of the recent reforming ideas in model compression is the Lottery Ticket Hypothesis~\cite{frankle2018lottery}. Morcos et al.~\cite{morcos2019one} showed successful results of model compression by generalized lottery ticket hypothesis across different benchmark datasets and popular DNN architectures. Yu et al.~\cite{yu2019universally} explained a family of possible slimmable architectures by using a variable layer width switch, based on the batch-normalizaton layer. 

\subsection{Weight Decomposition}
The idea of decomposition is to fragment a really large weight matrix (or tensor) into a set of linear sequence of smaller tensors, such that maximum information is retained. Denton et al.~\cite{denton2014exploiting} proposed singular value decomposition (SVD) of the original weight tensor to find the orthogonal bases. Jaderberg et al.~\cite{jaderberg2014speeding} built a low-rank filter-bank approximation of the convolutional layer, to achieve upto 4.5x speedup and compression. Lebedev et al.~\cite{lebedev2014speeding} used the popular canonical polyadic decomposition (CP) to achieve layer compression. Yu et al.~\cite{yu2017compressing} proposed a SVD-free greedy alternative for generalized bilateral decomposition (GreBdec) of the convolutional layer. Kim et al.~\cite{kim2015compression} proposed an iterative method of Tucker based decomposition and fine-tuning to regain the original accuracy. Much recently, Li et al.~\cite{li2020group} proposed a single formulation to easily switch between channel pruning and weight decomposition, by applying group sparsity across the columns or the rows of the weight tensor, respectively. 

There are some inherent challenges with directly consuming some of the existing solutions on model optimization. Firstly, it is very difficult to measure the maximum percentage of achievable compression, such that the accuracy does not drop below an admissible threshold. Ye et al.~\cite{ye2019adversarial} discuss these different challenges as a trade-off between model robustness and model compression. Secondly, the computational and resource requirements for model distillation and architecture search are very high. Especially, Liu et al.~\cite{liu2018rethinking} argued that it is more valuable to search for the pruned architecture shape instead of pruning the unimportant weight values and channels. Thirdly, it is not trivial to identify the rank of the low-rank approximation of the decomposable tensors. 
 




\section{System Architecture and Design}

\begin{figure*}[!ht]
	\includegraphics[width=.98\textwidth]{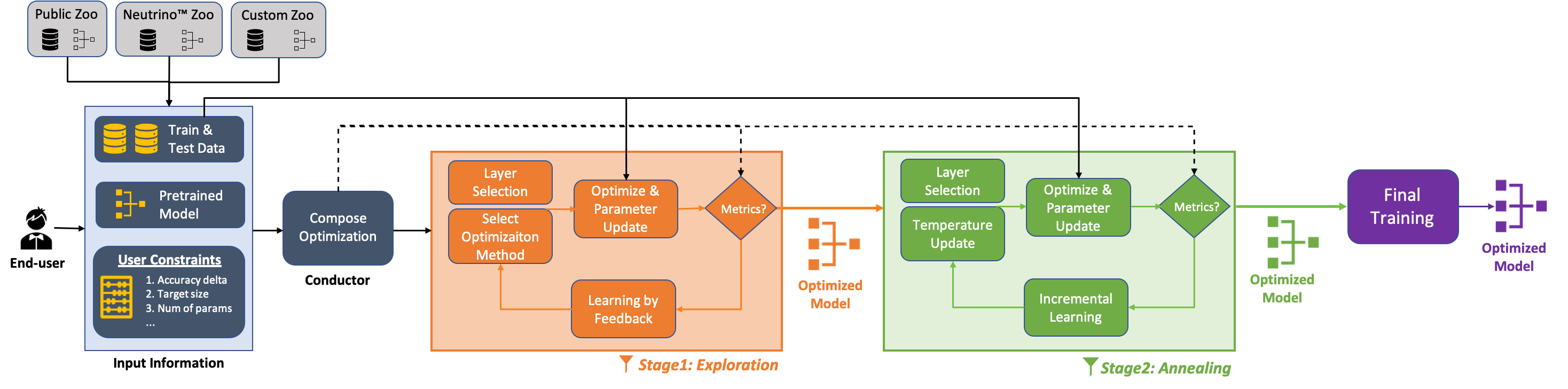}
	\caption{An overview of the architecture design highlighting the key components of the Neutrino framework.}
	\label{fig:overview2}
\end{figure*}

In this section, we describe the high-level solution architecture of Neutrino framework which contains four important components: (i) Neutrino Zoo, (ii) conductor, (iii) high-level coarse compression by exploration, and (iv) fine-grained aggressive compression by annealing. We focus on the system design from the end-users' usability perspective.
In this paper, we restrict the scope to optimizing convolutional neural networks (CNN) models for classification and object detection applications.

\subsection{Neutrino Zoo}

The end-user provides the following inputs to the framework: (a) a pre-trained model, $M$, (ii) the actual train-test data split used to train the model, ${DL_{train}}$ and $DL_{test}$, and (iii) a set of constraints or requirements to guide the optimization. The data pre-processing and data preparation steps performed during the original model training has to be reproduced in the provided data loaders. The pre-trained model and data loaders could be borrowed from any public github repository or any custom variant designed by the end-user. However, to ease the use of the end-user, a collection of popular DNN architectures with trained weights on different benchmark datasets are provided as Neutrino Zoo. The zoo consists of various classification and object detection datasets such as: MNIST, CIFAR10, CIFAR100, VWW, ImageNet, ImageNet10 (a 10-class subset of ImageNet), ImageNet16 (a 16-class subset of ImageNet), VOC2007, VOC2012, and COCO2017. Also, over $20$ trained DNN models are available including variants of ResNet, VGG, MobileNet, Inception, DenseNet, ShuffleNet, MLP, SSD with VGG/ MobileNet backbones, and YOLO-v3. The availability of the Neutrino Zoo allows the end-users to easily and quickly use the framework for transfer learning.

\subsection{Conductor}

The purpose of the conductor is to collect all the provided inputs, understand the given requirements, and orchestrate the entire optimization pipeline, accordingly.
The constraints to guide the optimization are provided by the end-user and the conductor automatically orchestrates the pipeline, by additionally inferring the model and data properties. Some of the common configurable parameters are:
\begin{enumerate}
    \item \textbf{delta:} The acceptable tolerance of accuracy drop with respect to the original model, for example, 1\%.
    \item \textbf{stage:} The two different stages of compression, while stage 1 is less intensive compression requiring fewer computational resources, stage 2 provides more aggressive compression using more resources and time.
    \item \textbf{device:} Perform the entire optimization and model inference in either CPU, GPU, or multi-GPU (distributed GPU environment).
    \item \textbf{modularity}: The end-user can customize multiple parts of the optimization process for Neutrino to adapt over more complex scenarios. Support for customization goes beyond vanilla classification, including specialized dataloader, custom backpropagation optimizer, and intricate loss function that their native library implementation allows.
\end{enumerate}

Let the pre-trained model has $N$ optimizable layers: $\{L_1, L_2, ..., L_N\}$. In a typical CNN model, the convolutional layers and the fully connected layers are optimizable while the rest of the layers are ignored from the optimization process. The conductor analyzes the data size, number of output classes, model architecture, and optimization criteria, \textit{delta}, and produces a binary composed list, $CL=\{c_1, c_2, ..., c_N\}$, where $c_i \in \{0, 1\}$. The conductor identifies the subset of optimizable layers that needs to be optimized, marked as $1$, and the layers that has to be frozen throughout the process, marked as $0$. This information is passed forward to the exploration stage, where the subset marked as $1$ is optimized.

\subsection{Stage 1: Exploration}

In a convolutional neural network, every optimizable layer, $L_i$ projects the input data into different dimensional outputs, as follows, 
\begin{equation}
    L_i \Longrightarrow \tilde{y}_i = f(W_i \otimes X_i)
\end{equation}
where $W_i$ is the kernel parameters of the layer, $X_i$ is the input, $\tilde{y}_i$ is the output, and $f$ is usually a non-linear activation function such as ReLU, sigmoid, or tanh, and $\otimes$ is the projection function.

\subsection{Transforming Layers:}
A transformation function is applied to every optimizable layer of the Convolutional Neural Network. This transformation function is designed to ensure that it approximates the original projection, $L_i$ while reducing the number of parameters of the layer. 

An n-D tensor can be viewed as a linear combination of multiple $1$-dimensional vectors using variable-separable method. For a layer having a parameters as a 4-D tensor of the shape \textit{[width $\times$ height $\times$ in\_shape $\times$ out\_shape]}, the following transformation function is applied, 
\begin{multline}
    W_i(w, h, in, out) = \\ 
    \sum_{r=1}^{R} W_i^{(1)}(w; r)W_i^{(2)}(h; r)W_i^{(3)}(in; r)W_i^{(4)}(out; r)
\end{multline}
with a canonical small-size $r$. During the forward pass, the transformation function of $L_i$ is performed as follows:
\begin{equation}
\begin{split}
    f(W_i \otimes X_i)  \Longrightarrow f(W_i^{(1)}W_i^{(2)}W_i^{(3)}W_i^{(4)} \otimes X_i) \\
 \Longrightarrow  
 \sum_{r=1}^{R}w_i^{(4)} \otimes 
    \bigg( 
        \sum_{a}w_i^{(1)} \otimes 
        \bigg( 
            \sum_{b}w_i^{(2)} \otimes  \\
            \bigg(  \sum_{c=1}^{in} w_i^{(3)} * x_i(a, b, c)  \bigg) \bigg) \bigg) 
\end{split}
\end{equation}

\noindent This transformation function reduces the number of layer parameters from \textit{(w * h * in * out)} to \textit{small\_size* (w + h + in + out)}.

For a layer for which $W_i$ is a 2-D matrix of the shape \textit{[in\_shape $\times$ out\_shape]}, the transformation function is designed, as follows,
\begin{equation}
    W_i(in, out) = \sum_{r=1}^{R} W_i^{(1)}(in, r)W_i^{(2)}(r, r)W_i^{(3)}(out, r)
\end{equation}
where, $r$ is the near-optimal small-size approximation of the original matrix. Thus, the layer's forward pass $L_i$ is replaced as follows, 
\begin{equation}
\begin{split}
    f(W_i \otimes X_i) \Longrightarrow f((W_i^{(1)}W_i^{(2)}W_i^{(3)})\otimes X_i) \\ \Longrightarrow f(W_i^{(1)}(W_i^{(1)} W_i^{(1)} \otimes X_i))
    \end{split}
\end{equation}

\noindent This reduces the overall number of parameters of $L_i$ from \textit{(in*out)} to \textit{small\_size * (in + out)}.

The challenge is to find an ideal small-size approximation, $r$, that produces good compression retaining the robustness of the model.
When the near-optimal small-size is equal to the actual size of the weight tensor, $r=small\_size(W_i)$, there is an over-approximation of the transformation with very low compression. A very small size, $r\xrightarrow[]{}0$, produces a high compression, however, with a lossy reconstruction of the transformation. The exploration stage searches for the near-optimal $r$, a lower size approximation of the tensor, $W_i$, such that there is minimal loss of the transformation function of the layer, $L_i$.

During the exploration stage, the composed list $CL=\{c_1, c_2, ..., c_N\}$ is updated, where Neutrino selects different transformation functions for different convolutional and fully connected dense layers. 
The entire model is optimized by the designed composition and the accuracy is regained by performing fine-tuning. The fine-tuning is performed using the same train-test data split used while pre-training the original model. The conductor checks if the optimized model adheres to the termination requirements as provided by the end-user, and if not, the composition list is updated and the next round of optimization is performed.

\subsection{Stage 2: Annealing}
Stage 2 optimization aims to perform aggressive compression and to obtain the maximum possible compression in the required tolerance of accuracy. For example, if the \textit{delta} of accuracy is $1\%$, and stage 1 produces a $4x$ compression with an accuracy drop of $0.6\%$, the aim of stage 2 is to further the compression with the \textit{delta} going as close as possible to $1\%$. 
In stage 2, the composed list $CL=\{c_1, c_2, ..., c_N\}$ of different layers is frozen, while the extent of optimization for each layer is increased. Annealing is a metaheuristic approach to approximate global optimization. By increasing the temperature of each layer, the overall energy of the model is preserved while finding a smaller size, $r$ that better approximates the global optima.

The entire pipeline of Neutrino  framework could be executed in a distributed multi-GPU environment, to speed-up the time required for optimizing the model. To achieve this, Uber's Horovod\footnote{\url{https://eng.uber.com/horovod/}}~\cite{sergeev2018horovod} an open-source library is reused. Horovod supports different backend libraries including PyTorch and Tensorflow, and is easy to use and integrate.

\begin{table*}[!ht]
\begin{tabular}{|p{2.5cm}|p{1.5cm}|p{1.5cm}|p{1.5cm}|p{1.5cm}|p{1.5cm}|p{2.2cm}|p{1.6cm}|}
\hline
\multicolumn{1}{|l|}{\textbf{Architecture}}  & \textbf{Model} & \textbf{Accuracy (\%)} & \textbf{Size (MB)} & \textbf{MACs (Billions)} & \textbf{\#Params (Millions)} & \textbf{Memory Footprint (MB)} & \textbf{Execution Time (ms)} \\ \hline
\multirow{4}{*}{\textbf{Resnet18}}           & Original        & 76.8295           & 42.8014            & 0.5567                   & 11.2201                      & 48.4389                       & 0.0594                \\ \cline{2-8} 
     & Stage1         &       76.7871            &      7.5261              &     0.1824                     &           1.9729                   &  15.3928                             &0.0494                            \\ \cline{2-8} 
     & Stage2         &   75.8008                &  3.4695                  &   0.0790                        &         0.9095                    &   10.3965                            &   0.0376                            \\ \cline{2-8} 
     					
     & \textbf{Enh}            &   \textbf{-0.9300}                &     \textbf{12.34x}             &  \textbf{7.05x}                        &   \textbf{12.34x}                           &   \textbf{4.66x}                            &  \textbf{1.58x}                             \\ \hline
\multirow{4}{*}{\textbf{Resnet50}}           & Original        & 78.0657           & 90.4284            & 1.3049                   & 23.7053                      & 123.5033                      & 3.9926                 \\ \cline{2-8} 
     & Stage1         &   78.7402                &    25.5877                &  0.6852                        &   6.7077                           &         65.2365                      & 0.2444                            \\ \cline{2-8} 
     & Stage2         &    77.1680               &   8.4982                 &   0.2067                       &   2.2278                           &     43.7232                          &  0.1772                           \\ \cline{2-8} 
     & \textbf{Enh}            &   \textbf{-0.9400}                &  \textbf{10.64x}                  &  \textbf{6.31x}                        &   \textbf{10.64x}                           &  \textbf{2.82x}                             &  \textbf{1.49x}                           \\ \hline
\multirow{4}{*}{\textbf{VGG19}}              & Original        & 72.3794           & 76.6246            & 0.3995                   & 20.0867                      & 80.2270                       & 1.4238                    \\ \cline{2-8} 
     & Stage1         &     71.5918 &	3.3216 &	0.0631 &	0.8707	& 7.5440 &	0.0278                         \\ \cline{2-8} 
     & Stage2         & 71.6602	 & 2.6226 &	0.0479	& 0.6875  &	6.7399 &	0.0263                         \\ \cline{2-8} 
     & \textbf{Enh}            &      \textbf{-0.8300} &	\textbf{29.22x} &	\textbf{8.34x} &	\textbf{29.22x} &	\textbf{11.90x}	& \textbf{1.67x}                       \\ \hline
     
\multirow{4}{*}{\textbf{DenseNet121}}        & Original        & 78.4612           & 26.8881            & 0.8982                   & 7.0485                       & 66.1506                       & 10.7240          \\ \cline{2-8} 
     & Stage1         &  79.0348 &	15.7624	& 0.5477 &	4.132 &	61.8052	& 0.2814                        \\ \cline{2-8} 
     & Stage2         &   77.8085 &	6.4246 &	0.1917 &	1.6842 &	48.3280	& 0.2372                        \\ \cline{2-8} 
     & \textbf{Enh}            &  \textbf{-0.6500}	& \textbf{4.19x}	& \textbf{4.69x}	& \textbf{4.19x}	& \textbf{1.37x} &	\textbf{1.17x}                               \\ \hline
\multirow{4}{*}{\textbf{GoogleNet}}          & Original        & 79.3513           & 23.8743            & 1.5341                   & 6.2585                       & 64.5977                       & 5.7186          \\ \cline{2-8} 
     & Stage1         &      79.4922 &	12.6389 &	0.8606 &	3.3132	& 62.1568	& 0.2856                       \\ \cline{2-8} 
     & Stage2         &   78.8086 &	6.1083 &	0.386 &	1.6013 &	51.3652 &	0.2188                    \\ \cline{2-8} 
     & \textbf{Enh}            &   \textbf{-0.4900}	& \textbf{3.91x}	& \textbf{3.97x}	& \textbf{3.91x}	& \textbf{1.26x} &	\textbf{1.28x}                    \\ \hline
\multirow{4}{*}{\textbf{Mobilenet v1}}       & Original        & 66.8414           & 12.6246            & 0.0473                   & 3.3095                       & 16.6215                       & 1.8147     \\ \cline{2-8} 
     & Stage1         &       66.4355 &	6.4211 &	0.0286 &	1.6833	& 10.5500 &	0.0306                       \\ \cline{2-8} 
     & Stage2         &   66.6211 &	3.2878 &	0.017 &	0.8619 & 	7.3447 &	0.0286                     \\ \cline{2-8} 
     & \textbf{Enh}            &     \textbf{-0.4000} &	\textbf{3.84x} &	\textbf{2.78x} &	\textbf{3.84x} &	\textbf{2.26x} &	\textbf{1.13x}                     \\ \hline
\multirow{4}{*}{\textbf{shufflenet\_v2\_1\_0}}       & Original        & 69.9805	& 5.1731 &	0.0462 &	1.3561 &	12.3418 & 0.0357          \\ \cline{2-8} 
     & Stage1         &   68.9844                &   3.2792                 &  0.0285                        &   0.8596                           &     10.8947                          &    0.0361                    \\ \cline{2-8} 
     & Stage2         &   69.3262                &  1.9315                  &  0.016                        &    0.5063                          &   9.3258                            &   0.0344                       \\ \cline{2-8} 
     & \textbf{Enh}            &  \textbf{-0.6500} &	\textbf{2.68x}  &	\textbf{2.89x}  &	\textbf{2.68x}  &	\textbf{1.32x} &	\textbf{1.04x}                      \\ \hline

\end{tabular}
\caption{Performance on different metrics obtained after multiple stages of optimization on the CIFAR-100 dataset, validating the enhancement (Enh) obtained using the proposed framework. All the results are computed for an input delta accuracy of $1\%$.}
\label{tab:cifar100}
\end{table*}

\begin{table*}[!ht]
\begin{tabular}{|p{2.5cm}|p{1.5cm}|p{1.5cm}|p{1.5cm}|p{1.5cm}|p{1.5cm}|p{2.2cm}|p{1.6cm}|}
\hline
\multicolumn{1}{|l|}{\textbf{Dataset}}  & \textbf{Model} & \textbf{Accuracy (\%)} & \textbf{Size (MB)} & \textbf{MACs (Billions)} & \textbf{\#Params (Millions)} & \textbf{Memory Footprint (MB)} & \textbf{Execution Time (ms)} \\ \hline
\multirow{4}{*}{\textbf{Imagenet16}}           & Original        & 94.4970	& 42.6663	& 1.8217  &	11.1847	& 74.6332  &	0.2158
          \\ \cline{2-8} 
     & Stage1         &     93.8179  &	3.3724  &	0.5155  &	0.8840	&  41.0819	& 0.1606                      \\ \cline{2-8} 
     & Stage2         &  93.6220  &	1.8220 &	0.3206 &	0.4776 &	37.4608	& 0.1341                       \\ \cline{2-8} 
     & \textbf{Enh}            &   \textbf{-0.8800}	& \textbf{23.42x}	 & \textbf{5.68x}	& \textbf{23.42x} &	\textbf{1.99x} &	\textbf{1.61x}                      \\ \hline

\multirow{4}{*}{\textbf{VWW}}           & Original             & 93.5995	& 42.6389 &	1.8217 &	11.1775	& 74.6057 &	0.2149    \\ \cline{2-8} 
     & Stage1         &  93.8179 &	3.3524 &	0.4014 &	0.8788 &	39.8382	& 0.1445                        \\ \cline{2-8} 
     & Stage2         &    92.6220	& 1.8309 &	0.2672 &	0.4800 &	36.6682	& 0.1296                     \\ \cline{2-8} 
     & \textbf{Enh}            &   \textbf{-0.9800}	& \textbf{23.29x}	& \textbf{6.82x}	& \textbf{23.29x} & \textbf{2.03x} &	\textbf{1.66x}                       \\ \hline

\end{tabular}
\caption{Performance of the ResNet18 model against multiple large scale datasets, validating the enhancement (Enh) obtained using the proposed framework. All the results are computed for an input delta accuracy of $1\%$.}
\label{tab:vww}
\end{table*}

\section{Experimental Results and Analysis}
In this section, we experimentally showcase the performance of the Neutrino in optimizing different CNN models. The different metrics used to evaluate the extent of optimization are explained, along with the experimental protocol.

\subsection{Metrics}
There are different metrics used to measure the amount of optimization and performance of Neutrino, as follows:
\begin{enumerate}
    \item \textbf{Accuracy:} The top-1 accuracy ($\%$) or the equivalent performance objective of the model is measured. Successful optimization retains the accuracy of the original model.
    \item \textbf{Model Size:} The disk size (MB) occupied by the trainable parameters of the model. Lower model size enables models to be deployed into devices with memory constraints.
    \item \textbf{MACs:} The computational complexity of the model is measured by the number (billions) of Multiply-Accumulate Operation (MAC) computed across the layers of the model. The lower the number of MACs, the better optimized is the model.
    \item \textbf{Number of Parameters:} Total number (millions) of trainable parameters (weights and biases) in the model. Optimization aims to reduce the number of parameters.
    \item \textbf{Memory Footprint:} The total memory (MB) required to perform the inference on a batch of data, including the memory required by the trainable parameters and the layer activations. A lower memory footprint is achieved by better optimization.
    \item \textbf{Execution Time:} The time (ms) required to perform forward pass on a batch of data. Optimized models have a lower execution time.
\end{enumerate}

\subsection{Experimental Protocol}
The results are shown using several different popular CNN models against three different benchmark datasets: CIFAR-100, ImagetNet16, and Visual Wake Words (VWW). All the optimization experiments are run with an end-user requirement of accuracy \textit{delta} of $1\%$.  The experiments are executed with a mini-batch size of $1024$, while the metrics are normalized for a mini-batch size of $1$. All the experiments are run on four parallel GPU, using horovod, and each GPU is a Tesla V100 SXM2 with 32GB memory.  The standard train-test split is used for the experiments. The images are $z$-normalized with global mean and variance computed from the training data. To make the training more robust, data augmentation is performed using random cropping of $80\%$ with resizing and random horizontal flip.

\subsection{Result Analysis}

The optimization results obtained using Neutrino across different popular CNN models on CIFAR-100 dataset are shown in Table~\ref{tab:cifar100} and the results of ResNet-18 architecture on different large scale vision datasets are shown in Table~\ref{tab:vww}. From Table~\ref{tab:cifar100}, it can be observed that the difference between the original and the final optimized model is less than $1\%$, based on the provided \textit{delta} requirement. Depending on the architecture of the original model, it can be observed that the model size could be compressed anywhere between $\sim$3x to $\sim$30x. VGG19 is known to be one of the highly overparameterized CNN models, and as expected, achieved a $29.22$x reduction in the number of parameters with almost $\sim$12x compression in the overall memory footprint and $\sim$8.3x reduction in computation complexity. The resulting VGG19 model occupies only 2.6MB as compared to the original model requiring 76.6MB. Mobilenet architectures are specifically designed to be lightweight with low computational cost, and even in Mobilenet v1, Neutrino achieved a size compression of $2.78$x with only $0.4\%$ reduction in accuracy. In a GPU environment, a speedup of around $\sim$1.5x is observed. This could significantly impact the inference time on the model, especially on the edge devices, and also the fine-tuning time required in future versions of production releases. The performance of Neutrino on large scale vision datasets produces around $\sim$23.5x compression of ResNet18 on Imagenet16 and VWW datasets. The optimized model requires only 1.8MB as compared to 42.6MB required by the original model. There is more than $\sim$1.6x in speedup with $5.7 - 6.8$x reduction in the computational complexity of the model. Crucially, it can be observed that Stage 2 compresses the model at least $2$x more than Stage 1 compression.

\begin{figure}[!t]
	\includegraphics[width=.48\textwidth]{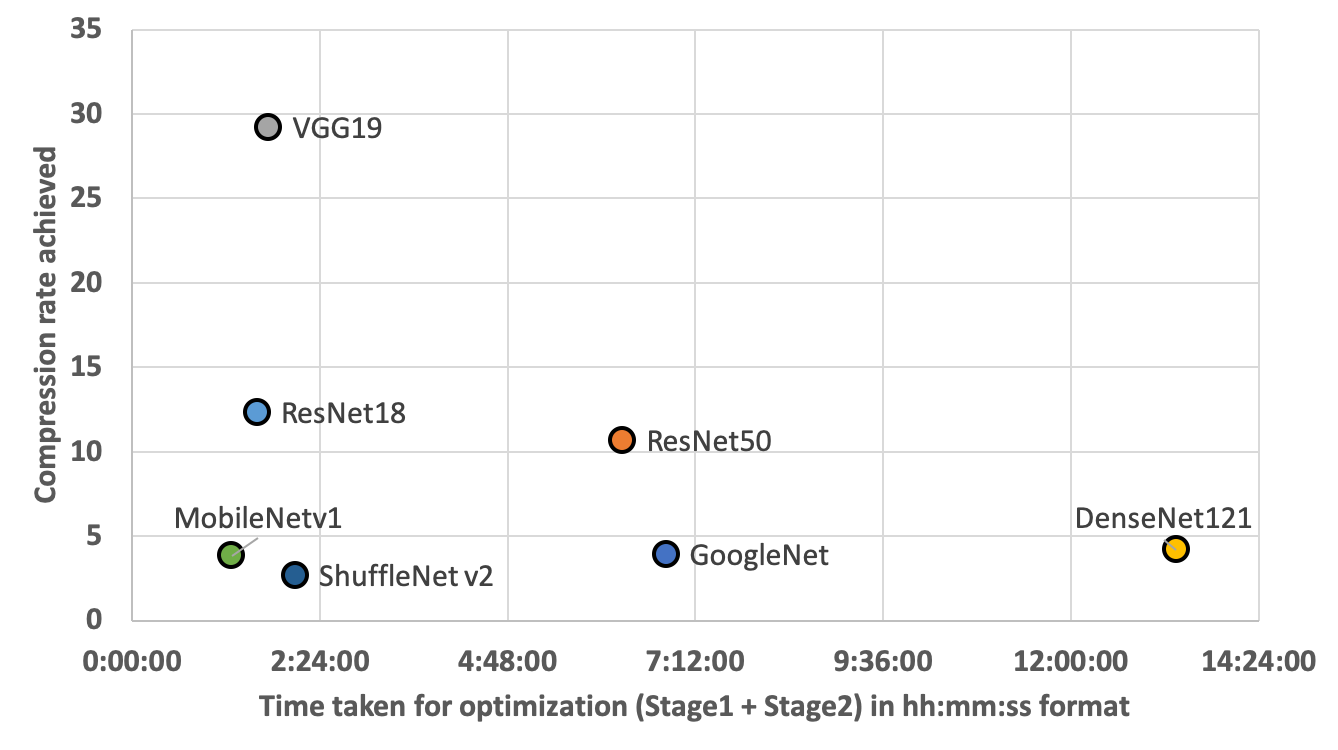}
	\caption{The total time taken for optimizing various models and the amount of compression achieved against CIFAR-100 dataset, using Neutrino framework.}
	\label{fig:time}
\end{figure}

\begin{figure}[!h]
	\includegraphics[width=.48\textwidth]{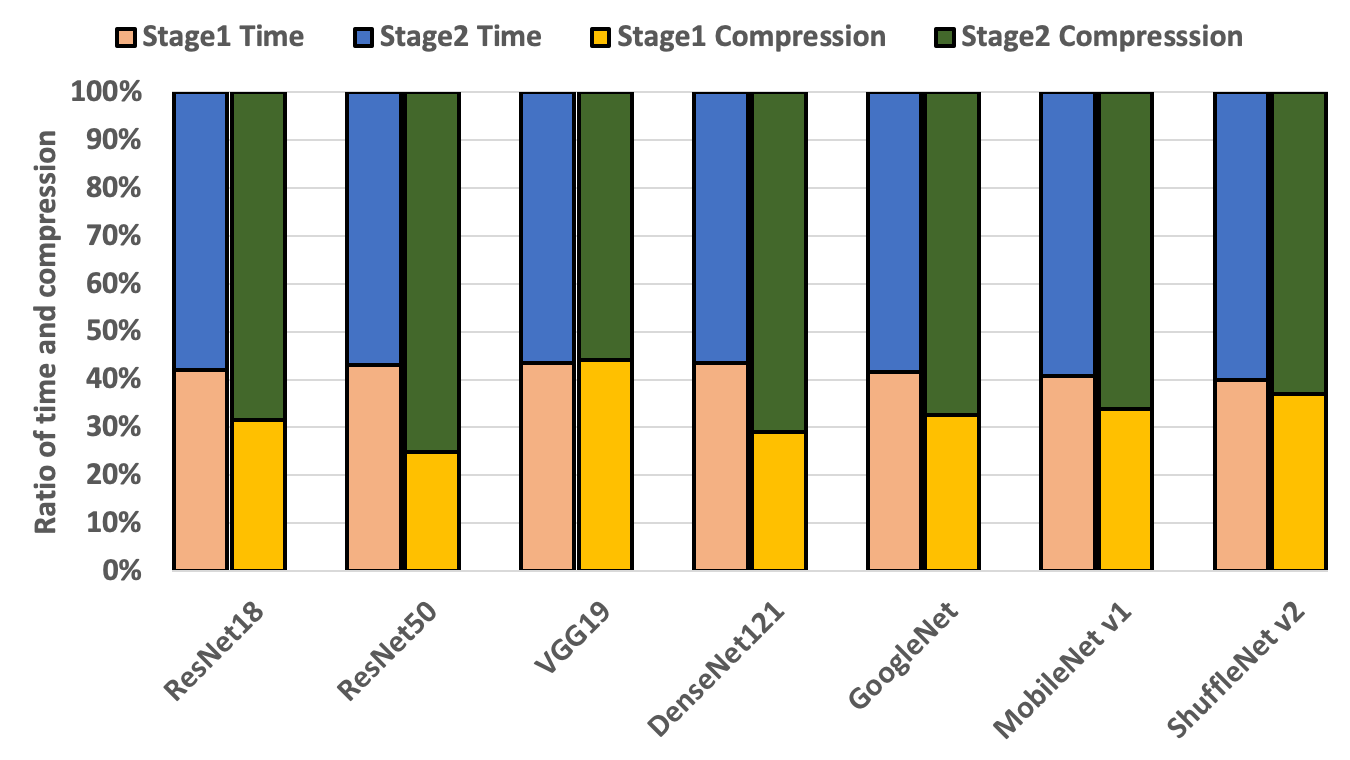}
	\caption{The proportion of time taken in optimization and the amount of compression, between Stage 1 and Stage 2 of optimization in the Neutrino framework.}
	\label{fig:results}
\end{figure}

\subsection{Time Taken for Optimization}

The overall time taken for optimization by Neutrino, including Stage 1 and Stage 2, is shown in Figure~\ref{fig:time}. It can be observed that most of the models could be optimized in less than $\sim$2 hours. While complex architectures, with longer training times, such as Resnet50 and DenseNet121 take around $\sim$6 hours and $\sim$13 hours for optimization, respectively. 
The comparison between the time taken for stage 1 and stage 2 compression is visually shown in Figure~\ref{fig:results}. It can be observed that almost $60-70\%$ of the overall optimization is achieved in Stage 2, while Stage 1 consumes less than $\sim$40$\%$ of the overall time required. This differentiation acts as a key feature of Neutrino, where end-users who need quick optimization with less resource consumption can choose Stage 1, while those needing aggressive optimization can choose Stage 2 optimization.  

It can be experimentally observed that Neutrino could be generalized across all kinds of CNN architectures and all scales of datasets with varying number of classes. Neutrino uniformly provides high metrics of optimization across all these datasets.

\begin{table*}[!ht]
\begin{tabular}{|p{1cm}|p{1.5cm}|p{1.4cm}|p{1.4cm}|l|l|l|l|p{1.1cm}|}
\hline
\textbf{Client}          & \textbf{Model}                     & \textbf{Dataset}                     & \textbf{Method}              & \textbf{Acc. (\%)} & \textbf{\#Params (M)} & \textbf{Size (bytes)}  & \textbf{FLOPS (M)} & \textbf{Time (ms)} \\ \hline
\multirow{3}{*}{Andes}   & \multirow{3}{*}{\parbox{1.1cm}{Mobile- NetV1}}       & \multirow{3}{*}{VWW}                 & Original                     & 88.1             & 3.2085                     & 12,836,104                &   105.7                                         &            -                  \\ \cline{4-9} 
                         &                                    &                                      & \textbf{Neutrino}            & \textbf{87.6}    & \textbf{\parbox{1cm}{0.1900 (16.88\%)}}    & \textbf{\parbox{1cm}{188,000 (68x)}}     &  \textbf{\parbox{1cm}{24.6}}                       &  -                                                  \\ \cline{4-9} 
                         &                                    &                                      & TFLM                         & 84.0              & \parbox{1cm}{0.2134 (15.03\%)}           & \parbox{1cm}{860,000 (14.9x)}            &  -                       &    -                                                \\ \hline
\multirow{4}{*}{Prod\#1} & \multirow{4}{*}{\parbox{1.6cm}{Mobile- NetV2-0.35x}} & \multirow{4}{*}{\parbox{1.2cm}{Imagenet Small}}      & Original                     & 80.9           & 0.4093                   & 1,637,076                  & 66.50                                  & 1.64                         \\ \cline{4-9} 
     &                                    &                                      & \textbf{Neutrino}            & \textbf{80.4}  & \textbf{\parbox{1cm}{0.1688 (58.76\%)}}  & \textbf{\parbox{1cm}{675,200 (2.4x)}}    & \textbf{50.90}            & 1.87                         \\ \cline{4-9} 
     &                                    &                                      & Intel Distiller              & 80.4           & \parbox{1cm}{0.2562 (37.41\%)}          & \parbox{1cm}{1,637,076 (1x)}             & 66.50    & 1.59                         \\ \cline{4-9} 
     &                                    &                                      & Microsoft NNI                & 77.4            & \parbox{1cm}{0.2851 (30.35\%) }        & \parbox{1cm}{1,140,208 (1.43x)}          & 52.80                & 2.22                         \\ \hline
\multirow{3}{*}{Prod\#2} & \multirow{3}{*}{\parbox{1.1cm}{Mobile- NetV2- 1.0x}}  & \multirow{3}{*}{\parbox{1.2cm}{Imagenet Small}}      & Original                     & 90.9           & 2.2367                   & 8,951,804                  & 312.8                 & 4.14                         \\ \cline{4-9} 
     &                                    &                                      & \textbf{Neutrino}            & \textbf{82.0}  & \textbf{\parbox{1cm}{0.4254 (80.98\%)}} & \textbf{\parbox{1cm}{1,701,864 (5.26x)}} & \textbf{134.00}      & 4.2                          \\ \cline{4-9} 
     &                                    &                                      & Intel Distiller              & 82.0           & \parbox{1cm}{0.2983 (86.66\%)}         & \parbox{1cm}{8,951,804 (1x)}             & 312.86                               & 4.4                         \\ \hline
\multirow{2}{*}{Prod\#3} & \multirow{2}{*}{\parbox{1.1cm}{Mobile- NetV2- 0.35x}} & \multirow{2}{*}{\parbox{1.2cm}{Gesture Recognition}} & Original                     & 96.8           & 2.3630                      & 10,500,000                          & 559.60                   &                706                                  \\ \cline{4-9} 
         &                                    &                                      & \textbf{Neutrino}            & \textbf{96.8}  & \textbf{\parbox{1cm}{0.5525 (76.62\%)}}  & \textbf{\parbox{1cm}{2,199,200 (4.77x)}}                & \textbf{508.20}                            &  611                            \\ \hline

\multirow{2}{*}{Prod\#4} & \multirow{2}{*}{\parbox{1cm}{SSD300 (ResNet50)}}  & \multirow{2}{*}{COCO-10}             & Original                     &  0.438 (mAP)                 &  14.17                          & 56,734,728                 &  15.59                        &   3.98                                             \\ \cline{4-9} 
     &                                    &                                      & \textbf{Neutrino} & \textbf{0.433 (mAP) }         & \textbf{\parbox{1cm}{4.84 (2.93x)}}                   & \textbf{\parbox{1cm}{19,365,488 (2.93x)}} &  \textbf{\parbox{1cm}{5.254 }}                      &                2.76                                 \\ \hline
\end{tabular}
\caption{Results from different production applications and business use-cases of Neutrino framework. It can be observed that in many practical real-world applications Neutrino performs better than other competitive optimization frameworks. The results are computed across different hardware deployments. The names of certain clients and production environments are redacted for anonymization.}
\label{tab:business}
\end{table*}

\section{Business Impact}

Our blackbox optimization framework has been deployed into multiple real-world applications and has been consumed by different clients. From different chip manufacturers enabling edge deployment of DNN architectures, to a faster inference of computer vision models on the cloud, the Neutrino framework could cater to a wide variety of use-cases. Some of the key real-world use-cases, where Neutrino is currently deployed in production are:
\begin{itemize}
    \item \textbf{Smart Appliances:} More than 100 million home appliances currently use ARM on Raspberry Pi 4 with only 2GB memory. To enable on-device, AI-driven, automated gesture recognition, Neutrino is used to compress MobileNet variant architectures by almost $\sim$2.5x.
    \item \textbf{Person Detection:} An embedded system with a small camera which uses RISC-V CPU cores~\cite{waterman2011risc}, is used as a home assistant alarm, by doing person detection. To enable very large DNN architectures to be deployed on these CPU cores, Neutrino framework is used to achieve up to $\sim$68x compression.
    \item \textbf{Autonomous Driving:} To enable autonomous self-driving cars, it is needed to perform real-time object detection with a highly noisy background. A highly complex DNN architecture: SSD-300 with ResNet50 as the backbone is used to accomplish object detection. However, for this large DNN model to be deployed inside an NVIDIA Xavier GPU, Neutrino framework is used to achieve $\sim$3x compression, along with $\sim$3x speedup, and $\sim$3x in power reduction, with no reduction in accuracy. 
\end{itemize}

The results obtained from the real-world deployments across various use-cases are shown in Table~\ref{tab:business}. It can be observed from the results that across different production environments, use-cases, models, and datasets, the Neutrino can be generalized for successful compression of models. Depending on the application requirements, Neutrino produces anywhere between $\sim$2x to $\sim$68x compression, with less than $\sim$1\% accuracy reduction from the original model. Also, in the same production environments, Neutrino was compared with competitive optimization frameworks such as Microsoft's Neural Network Interface (NNI)\footnote{https://github.com/microsoft/nni}, Intel's Neural Network Distiller\footnote{https://github.com/NervanaSystems/distiller}, and Tensorflow Lite Micro\footnote{https://www.tensorflow.org/lite/microcontrollers}. It can be observed that Neutrino consistently outperforms the competitors by achieving higher compression with better accuracy. 
As a testimonial to the success and usability, Neutrino framework has received several accolades and media coverage, some of which are listed here:
\begin{itemize}
    \item Neutrino and the parent company Deeplite has been named the \textbf{AI 100}, one of the top 100 AI companies globally, by CB Insights\footnote{https://www.cbinsights.com/research/artificial-intelligence-top-startups/}. CB Insights platform annually chooses the list from a candidate set of more than $5000$ companies, with technical novelty being one of the primary criteria. Also Intel Capital, in their AI infrastructure stack landscape~\footnote{\url{https://www.intel.com/content/www/us/en/intel-capital/news/story.html?id=a0F1I00000BNTXPUA5}}, has identified Deeplite and Neutrino , as one of the few production-ready optimization frameworks available in the market today.
    \item In a joint partnership, Deeplite and Andes Technologies used Neutrino to deploy optimized DNN models on the first commercial RISC-V cores based on AndeStar V5 architecture\footnote{\url{https://www.prnewswire.com/news-releases/andes-technology-and-deeplite-inc-join-forces-to-deploy-highly-compact-deep-learning-models-into-daily-life-300972366.html}}. In a specific use-case, a MobileNet-v1 model trained on a Visual Wake Words (VWW) dataset was compressed from 13MB to only 688KB ($\sim$68 times compression) with less than 1\% drop in accuracy. According to Dr. Charlie Su, CTO and Executive VP of Andes Technology, \textit{``Deeplite has provided a solution that can be leveraged both internally within Andes as well as for our customers to bring deep learning on Andes RISC-V CPU cores to resource-limited devices at the edge."}
    \item Using Neutrino framework large DNN architectures are currently being optimized and also deployed in ARM microcontrollers\footnote{\url{https://community.arm.com/developer/ip-products/processors/b/ml-ip-blog/posts/unlocking-ai-on-arm-microcontrollers-with-deep-learning-model-optimization}}. In a specific-use case of low-power camera, the underlying ARM Cortex-M4 has a memory resource constraint of only 256KB on-chip memory. Automatically guided by the memory constraint, Neutrino compressed a 13MB large DNN architectures to only 144KB ($\sim$88 times compression) with less than 1.84\% accuracy drop as compared to the original model.
\end{itemize}

\noindent There is CI/CD based DevOps pipeline, with a monthly sprint delivering product enhancements, software patches, and bug fixes. There is a committed core team of eight technical developers (and growing fast) with diverse skills, to lead and support new features and new client deployments.

\section{Conclusion and Future Work}
In this paper, we proposed an easy-to-use blackbox framework for DNN model optimization, Neutrino. The framework is completely automated and could be used to optimize any convolutional neural network based architecture, with no human intervention. The end-user could provide the requirements of optimization such as target model size, or the tolerance drop in accuracy, and Neutrino  framework would produce the optimized model, according to the requirements. As an experimental validation, the performance of the proposed framework was shown against several benchmark datasets and popular architectures. Neutrino  is currently in production and is used by several clients for multiple use-cases such as smart appliances, autonomous driving, or person detection. The success of the framework in production, along with several testimonials, are showcased. Following the challenges presented in the first section for model optimization, Neutrino  is a robust and early solution that only scratches the surface. Therefore, some of the ongoing and future work has much potential to offer, such as being more target hardware aware and further improving compression and speed-up by using techniques.

{\small
    \bibliography{egbib}
}

\end{document}